# Multi-robot SLAM Multi-view Target Tracking based on Panoramic Vision in Irregular Environment

Wang Ruiqi, Ziqin Yuan, Guohua Chen
Beijing University Of Chemical Technology, Beijing, 100029,China

**Abstract:** In order to improve the precision of multi-robot SLAM multi-view target tracking process, a improved multi-robot SLAM multi-view target tracking algorithm based on panoramic vision in irregular environment was put forward, adding an correction factor to renew the existing Extended Kalman Filter (EKF) model, obtaining new coordinates X and Y after twice iterations. The paper has been accepted by *Computing and Visualization in Science* and this is a simplified version.

## 1. Methods:

### 1.1 Extended Klman filter model

The kalman filter system is assumed to be a linear system, but the actual robot motion model and the observation model are nonlinear. Therefore, extended kalman filters are usually adopted. The mapping and positioning based on EKF can be summarized as a circular, iterative estimation - calibration process. EKF algorithm has a unique feature in processing uncertain information, so EKF becomes the most widely used SLAM method. According to the above equations and assumptions, the measured value $x_k$ of kalman filter is the target position calculated from the state prediction vector. It refers to the specific position of the target point in the robot coordinate system, which can be expressed as(wk is the correction factor):

$$H^i(X_k) = \begin{bmatrix} x^i_{Rk} \\ y^i_{Rk} \end{bmatrix} = R^T \begin{bmatrix} x^i_{Gk} - x_k \\ y^i_{Gk} - y_k \end{bmatrix} + w^i_k \qquad (6)$$

Therefore, the system state equation and measurement equation are as follows:

$$\begin{cases} X_k = F_{k-1} X_{k-1} + u_{k-1} + v_{k-1} \\ Y_k = H_k X_k + w_k \end{cases} \qquad (7)$$

Where,

$$F_k = \frac{\partial F}{\partial X}\Big|_k, \; H_k = \begin{bmatrix} \frac{\partial H^1}{\partial X} & \cdots & \frac{\partial H^2}{\partial X} & \cdots & \frac{\partial H^n}{\partial X} \end{bmatrix}'\Big|_k. \qquad (8)$$

### 1.2 Renewal equation

EKF is used to estimate and update the system state. When a new feature is found during the motion of the robot, it is required to calculate the initial state marked by the new feature according to the observation vector z(k+1) of the new feature and the current state of the robot, and to update the state vector and covariance matrix P.

$$P = \begin{bmatrix} p_{rr} & p_{r1} & \cdots & p_{rn} \\ p_{1r} & p_{11} & \cdots & p_{1n} \\ \vdots & \vdots & \ddots & \vdots \\ p_{nr} & p_{n1} & \cdots & p_{nn} \end{bmatrix} \qquad (9)$$

In SLAM, the system state includes the estimation of the positions of the robot and the environment feature in the robot coordinate system, while the covariance matrix P represents the error of estimation. The information processing by the method of extended kalman filter is generally divided into prediction and update, and the information estimation by this method is unbiased estimation.

The kalman filter prediction equation is as follows:

$$\begin{cases} X_k^- = F_{k-1}X_{k-1} + u_k, \\ Y_k = H_k X_k^-, \\ P_k^- = F_{k-1}P_{k-1}F_{k-1}^T + Q_{k-1} \end{cases} \quad (10)$$

The update equation is as follows:

$$\begin{cases} K_k = P_k^- H_k^T (H_k P_k^- H_k^T + R_k)^{-1}, \\ X_k = X_K^- + K_k(Y_k - Z_k), \\ P_k = (I - K_k H_k)P_k^-, \end{cases} \quad (11)$$

Where, $Q$ refers to the state noise covariance; $R$ refers to the measurement noise covariance; $P$ the error covariance; and $K$ the gain matrix.

The prediction equation is used to predict the current state, while the error covariance matrix is used to obtain a prior estimate for the next moment. The new equation combines the prior estimate with the measured value to obtain a more accurate posteriori estimate.

**1.3 SIFT algorithm**

In 2004, Lowe D G summarized the existing feature detection methods based on invariant technology, and formally proposed a local feature descriptor for images based on scale space. It is called SIFT descriptor which remains invariant to image scaling, rotation, and even affine transformation, namely scale invariant feature transform. Under the SIFT algorithm, firstly, feature detection is carried out in the scale space, and the position and scale of the key point are determined. Secondly, the main direction of the field gradient of the key point is taken as the direction feature of the point, so as to realize the independence of the operator to scale and direction. The calculation equation of gradient and direction is as follows:

$$m(x,y) = \left( (L(x+1,y) - L(x-1,y))^2 + (L(x,y+1) - L(x,y-1))^2 \right)^{1/2} \quad (12)$$

$$\theta(x,y) = a\tan 2(L(x,y+1) - L(x,y-1))/(L(x+1,y) - L(x-1,y)) \quad (13)$$

Where, $(x,y)$ refers to the gradient magnitude and direction equation, and the scale used by $L$ is the scale of each key point. In the actual calculation, samples can be taken in the neighboring window centered on the key point, and the gradient direction of neighborhood pixels can be calculated by histogram. The peak value in the histogram represents the main direction of the neighborhood gradient at the key point, that is, the direction of the key point.

**2 Experimental analysis**

**2.1 Experiment setting**

(1) Experiment hardware: The processor CPU is i5-6400K 3.2GHz. Internal storage RAM is ddr4-2400GHz 8GB. The system is win10 flagship version. The simulation platform is Matlab2012a.

(2) Dataset: The PETS 2009 dataset is selected as the experiment object. Here is an example of dataset in Fig.3. Three sets are used in the scenarios, namely S2.L1, S2.L2 and S2.L3. The devices have four to seven different perspectives, from overlapping outdoor robot vision cameras. Each image has hundreds of frames. 10-74 pedestrians are captured at 7 frames per second. In different scenarios, there are different pedestrian densities, from S2.L1 (low) to S2.L3 (high). The benchmark dataset can also provide the internal and external parameters of each robot vision camera through TSAI's robot vision camera model. In the evaluation, the target position in the region of interest of each frame is given.

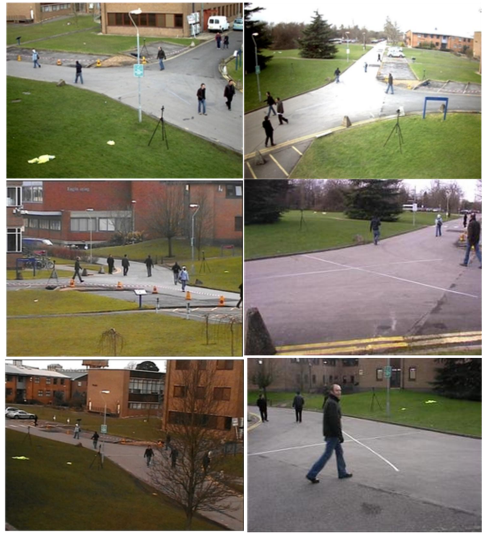

Fig.3 Example of PETS 2009 Dataset

In order to inspect the effect of the parameters and conditions in the proposed cost function on performance precision, a new dataset is constructed, namely pilsun dataset. Our dataset contains 333 frames from four overlapping cameras. Each of these cameras will capture 10 pedestrians. These cameras are densely distributed in a small indoor environment at 6 frames per second. The dataset also provides a TSAI camera model for all cameras and ground truth generated by the manually tagged trace results.

(3) Parameter setting: the comparison interval distance $L_c = 4$ frames ; the static feature point threshold $\delta_{min} = 0.05$ ; the maximum allowable 3D distance between continuous detections $\varepsilon_\Phi = 0.5\text{m}$ ; the change in the maximum allowable 3D height of a target in a frame $\varepsilon_h = 0.3\text{m}$ ; the size difference between adjacent windows $\omega_\delta(s_i) = 0.3 \times s_i$ ; the minimum 3D distance to avoid collision $\theta_s = 0.3\text{m}$ ; the maximum 3D distance detected at the same time $\varepsilon_{3D} = 2.5\text{m}$ ; the maximum 3D velocity of the target $v_{max} = 0.8\text{m}/s$ ; and the maximum allowable frame clearance on the track $\delta_a = 9$ frames .

(4) As quantitative evaluation indicators, multi-target tracking accuracy indicators MOTA and MOTP are used in the classification of events, activities and relationships, and they are also the most popular indicators in multi-robot SLAM multi-perspective target tracking.

## 2.2 Result analysis

(1) Parameter influence experiment. The operation parameter setting of the proposed method: $K_H = 1, 5, 10, 15, 20, 25, 30$ and $i_{bls}^{max} = 500, 1000, 2000$ . The influence of the maximum values of $K_H$ and $i_{bls}^{max}$ on the calculation accuracy and time is studied. Fig.4 shows the experiment on the influence of processing time and MOTA changing with parameters.

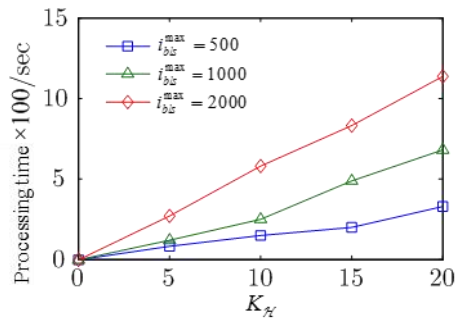

(a) Algorithm Processing Time

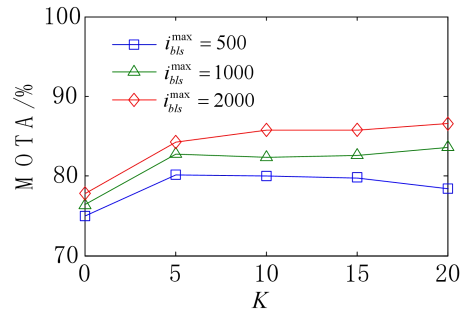

(b) MOTA Accuracy Indicator

Fig.4 Parameter Influence Experiment

As shown in Fig.4, the processing time also increases linearly as $i_{bls}^{max}$ increases. However, if the $i_{bls}^{max}$ is not high enough, the performance will not increase. The last figure in Fig.4 represents the gap between the maximum MOTA and minimum MOTA in each setting. According to this figure, the smaller maximum value of $i_{bls}^{max}$ leads to unstable performance, while MOTA with enough high maximum $i_{bls}^{max}$ gradually stabilizes with $K_H$.

### 3.Conclusions

In this paper, an online multi-robot SLAM multi-perspective target tracking algorithm based on MHT framework was proposed. By using this scheme, multi-target tracking problems with small track sets can be generated according to the global assumptions in the previous framework. The solution searching space of these sub-problems is much smaller than the original multi-target control algorithm, so the solving time can be greatly reduced. In addition, the proposed scheme also provides a better solution than using the full trajectory to solve the multi-target control problems. The initial solution and iterative times proposed are helpful to find the approximate optimal solution with fewer iterative times. In the next step, attention will be paid to the research on the convergence analysis of the algorithm, aiming to improve the theoretical basis of the algorithm.